\documentclass[]{jingdong}

\usepackage{amsfonts}
\usepackage{nicefrac}
\usepackage{enumitem}
\usepackage{pifont}

\usepackage{amsmath,amssymb}
\usepackage{amsthm}
\usepackage{algorithm}
\usepackage{algpseudocode}
\usepackage{colortbl}
\usepackage{wrapfig}
\usepackage{booktabs}
\usepackage{tabularx}
\usepackage{array}
\usepackage[table]{xcolor}
\usepackage[utf8]{inputenc}

\microtypesetup{expansion=false}
\graphicspath{{figures/}}
\definecolor{SkillSection}{HTML}{F3F6FA}
\definecolor{SkillTotal}{HTML}{E8EEF7}
\definecolor{SkillStripe}{HTML}{FAFAFA}
\definecolor{rubricblue}{HTML}{D85C55}

\newcommand{\catrule}{%
  \addlinespace[0.2em]
  \cmidrule(lr){1-5}
  \addlinespace[0.1em]
}

\title{SkillCoach: Self-Evolving Rubrics for 
Evaluating and Enhancing Agentic Skill-Use}

\author[1,\ddagger]{Jiayin Zhu}
\author[2]{Kelong Mao}
\author[2]{Yudong Guo}
\author[1]{Dengbo He}
\author[2]{\protect\\[+0.25em]Sulong Xu}
\author[2]{Simiu Gu}
\author[1,*,\dagger]{Yutao Yue}

\affiliation[1]{HKUST(GZ)}
\affiliation[2]{JD.COM}

\checkdata[Email]{\email{jzhu709@connect.hkust-gz.edu.cn}; \email{yutaoyue@hkust-gz.edu.cn}; \email{maokelong.1@jd.com}}

\abstract{Skills are becoming a reusable operational layer for LLM agents, encoding SOPs, domain rules, tool workflows, scripts, and validation routines. In realistic skill repositories, overlapping skills make reliable skill-use difficult. Final verifier success is too coarse for both evaluation and training, since an agent may pass through trial and error while selecting distractor skills, skipping required steps, composing workflows incorrectly or omitting final checks.
We introduce \textsc{SkillCoach}, a self-evolving rubric framework for evaluating and enhancing agentic skill-use. \textsc{SkillCoach} derives skill-grounded process rubrics from real rollouts and evaluates trajectories along four dimensions: skill selection, skill following, skill composition, and skill-grounded reflection. It keeps the external verifier as a separate outcome signal, allowing process quality to be distinguished from accidental task success. The evolved rubrics further serve as process supervision for selecting high-quality training trajectories.
Experiments show that evolved rubrics substantially improve evaluation quality, expose failures hidden by final accuracy, and provide stronger supervision signals than outcome-only filtering for enhancing agentic skill-use.}

\begin{document}

\maketitle

\setlength{\skip\footins}{12pt}

\begingroup
\renewcommand{\thefootnote}{}
\footnotetext{
\textsuperscript{\ensuremath{\ddagger}}This work was completed during Jiayin Zhu's internship at JD.COM.

\quad
\textsuperscript{\ensuremath{\dagger}}Corresponding author.\quad
\textsuperscript{\ensuremath{*}}Yutao Yue is also affiliated with Institute of Deep Perception Technology, JITRI, Wuxi, China.
}
\endgroup

\vspace{-2mm}

\section{Introduction}
Skills are emerging as a practical operational layer for LLM agents. They package domain procedures, tool workflows, scripts, and validation checks into reusable units of operational knowledge. In enterprise settings, agents repeatedly follow internal processes and use tools. As skill libraries grow, the challenge shifts from having skills to using them reliably: selecting the right skill, following steps, composing workflows, and reflecting on outputs.

Recent work has studied agent skills through benchmarks, generation, retrieval, evolution, and lifecycle management~\citep{skillbench2026,skilllearnbench2026,skillflow2026,sokagentskills2026}. Other systems improve the skill artifact itself through discovery, refinement, verification, or optimization~\citep{evoskill2026,skillopt2026}. These directions are valuable, but they often rest on an implicit assumption: once a skill is generated or retrieved, the agent can use it correctly. In real skill libraries, this assumption often breaks down. As libraries grow and skills encode increasingly detailed business logic, agents may miss required skills, choose wrong ones, skip key steps, use skills in the wrong order or forget final checks before submission.
\begin{wrapfigure}{r}{0.60\linewidth}
    \centering
    \vspace{-0.8em}
\includegraphics[width=0.9\linewidth]{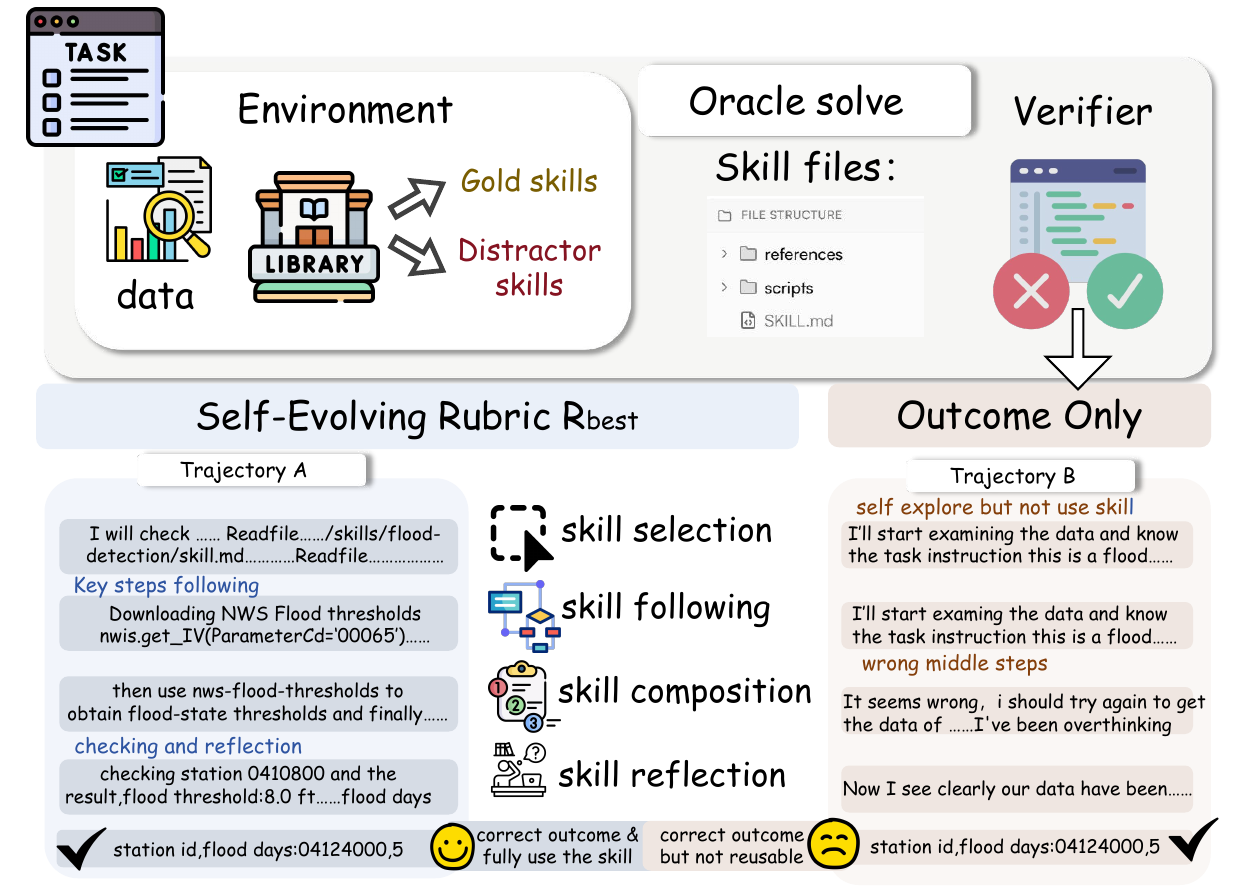}
    \caption{Motivating example of agentic skill-use diagnosis.}
    \label{fig:motivation}
    \vspace{-1.0em}
\end{wrapfigure}
Final task success is an incomplete signal for skill-using agents. Figure~\ref{fig:motivation} illustrates why outcome-only evaluation is insufficient: two trajectories can pass the verifier, while only one exhibits reusable skill-use behavior. The same issue also limits training. If all verifier-passing trajectories are treated as positive demonstrations, the agent may learn brittle behaviors such as selecting distractor skills, skipping required steps, relying on trial and error, or submitting outputs without skill-specified checks.

This motivates a process-level view of agentic skill-use. Recent agent evaluation work has moved from final outcomes to trajectory-level assessment~\citep{agentasjudge2024,agentprocessbench2026,toolprmbench2026}, and rubric-based methods show that task-adaptive criteria can provide actionable feedback~\citep{adarubric2026,autorubric2026}. However, existing work has not yet used the structure of skills themselves to define process supervision for both evaluation and training.

We propose \textsc{SkillCoach}, the first self-evolving rubric framework for evaluating and enhancing \textbf{agentic skill-use}. We define agentic skill-use as a trajectory-level meta-ability with four dimensions: \textbf{skill selection}, \textbf{skill following}, \textbf{skill composition}, and \textbf{skill-grounded reflection}. The external verifier is kept separate, which allows us to distinguish reliable skill-use behavior from accidental verifier passing. The evolved rubrics are then used in two ways: they diagnose skill-use failures, and they select high-quality trajectories for supervised skill-use training.

\textsc{SkillCoach} is designed for two deployment realities. First, we focus on \textbf{skill-dependent tasks}, where strong agents struggle without the relevant skill but improve when the gold skill is available. This shifts our focus from evolving skills to evolving rubrics that check how skills are used. Second, we evaluate agents in distractor-augmented skill libraries, where gold skills are mixed with distractor skills, approximating enterprise repositories with overlapping workflows.

For each task, \textsc{SkillCoach} derives an initial rubric from gold skills, scores real rollouts with observable evidence, and refines the rubric through validation-gated local patches. The final rubric serves as a process-level signal for both evaluation and improvement. It exposes where an agent fails to use skills correctly, and it filters verifier-passing trajectories into higher-quality demonstrations for supervised training. Experiments show that self-evolved rubrics improve evaluator quality, reveal skill-use failures hidden by final accuracy, and improve agentic skill-use training. These results suggest that skill-use is not only measurable, but also trainable through skill-grounded process supervision.

In summary, our contributions are:
\begin{itemize}

\item We first formulate agentic skill-use as a trajectory-level meta-ability spanning skill selection, skill following, skill composition, and skill-grounded reflection, kept separate from the external verifier signal.

\item We introduce \textsc{SkillCoach}, the first self-evolving rubric framework that calibrates task-level process rubrics from real rollouts—through evidence-grounded judging, local arbitration patches, and validation-gated updates—under a skill-dependent setting with distractor-augmented libraries that mirror enterprise repositories.

\item We show that the evolved rubrics serve a dual role: they improve evaluation quality and expose failures hidden by final accuracy, and as a supervision filter they select better trajectories than outcome-only SFT, with the meta-ability dimensions acting as dissociable supervision signals.

\end{itemize}

\section{Related Work}
\label{sec:related-work}

\paragraph{Agent skills as reusable execution abstractions.}
Agent skills package procedural knowledge, applicability conditions, workflows, and optional executable resources into reusable units for LLM agents. Recent surveys and systems increasingly treat such skills as deployment-time infrastructure rather than isolated prompt fragments~\citep{sokagentskills2026,cuaskill2026,declarativeskills2026}. This line clarifies how skills are represented, distributed, and invoked, but does not by itself determine whether an agent can use a provided library reliably. Our focus is therefore not merely whether a useful skill exists, but whether an agent recognizes when it applies, follows its prescribed procedure, and composes it with other skills when required.

\paragraph{Skill benchmarks and realistic skill use.}
\textit{SkillsBench} compares no-skill, curated-skill, and self-generated-skill settings, while \textit{SkillLearnBench} studies continual skill generation on verified skill-dependent tasks and evaluates both trajectory behavior and final outcomes~\citep{skillbench2026,skilllearnbench2026}. Studies closer to deployment examine retrieval and refinement over large real-world libraries, while \textit{SkillFlow} investigates lifelong skill discovery, patching, and transfer~\citep{liu2026agenticskillsworkwild,skillflow2026}. Collectively, these works show that high-quality skills can improve task success, yet their benefits become less reliable when agents must author skills, search large libraries, or reuse skills over time. However, their central evaluation targets remain skill utility, generation quality, retrieval, or transfer, rather than a library-grounded decomposition of whether an agent selected, faithfully followed, composed, and reflected on the intended skills.

\paragraph{Skill generation, evolution, and lifecycle management.}
A complementary line improves the skill artifact or the policy that consumes it. Failure-driven refinement, trajectory synthesis, validation-gated optimization, and evaluation-guided self-improvement update skills from execution evidence~\citep{evoskill2026,skillgen2026,skillopt2026,skillaxe2026}. Co-evolutionary verification and scientific-resource mining construct richer multi-file or domain-specific skill packages~\citep{CoEvoSkills2026,skillfoundry2026}. Unified lifecycle, reinforcement-learning, and memory-oriented approaches create, manage, and exploit expanding skill libraries for agent self-improvement~\citep{museautoskill2026,sage2025,skillrl2026,memskill2026}. These methods optimize which skills are available, how they are revised, or how a policy consumes them. Our work is complementary: it optimizes the evaluator and training signal used to determine whether observed rollouts exhibit reliable and reusable skill use, rather than merely successful task outcomes.

\paragraph{Trajectory-level and rubric-based evaluation.}
General agent evaluation has moved beyond final-answer accuracy toward full-trajectory assessment and process supervision. \textit{Agent-as-a-Judge} evaluates complete agent trajectories, while \textit{AgentProcessBench} and \textit{ToolPRMBench} target step-level process quality and process reward models for tool-using agents~\citep{agentasjudge2024,agentprocessbench2026,toolprmbench2026}. Rubric-based approaches such as \textit{AdaRubric} and \textit{Autorubric} further show that task-adaptive criteria can yield more specific and actionable judgments~\citep{adarubric2026,autorubric2026}. In the skill setting, \textit{Counterfactual Trace Auditing} demonstrates that pass-rate comparisons can conceal substantial behavioral effects introduced by skills~\citep{counterfactualtrace2026}. These methods motivate process-level supervision, but their criteria are not explicitly grounded in the structure of a concrete skill library. In particular, they do not jointly encode gold skills, distractors, required key steps, cross-skill dependencies, and skill-grounded reflection.

\begin{figure*}[t]

  \centering
  \includegraphics[width=\textwidth]{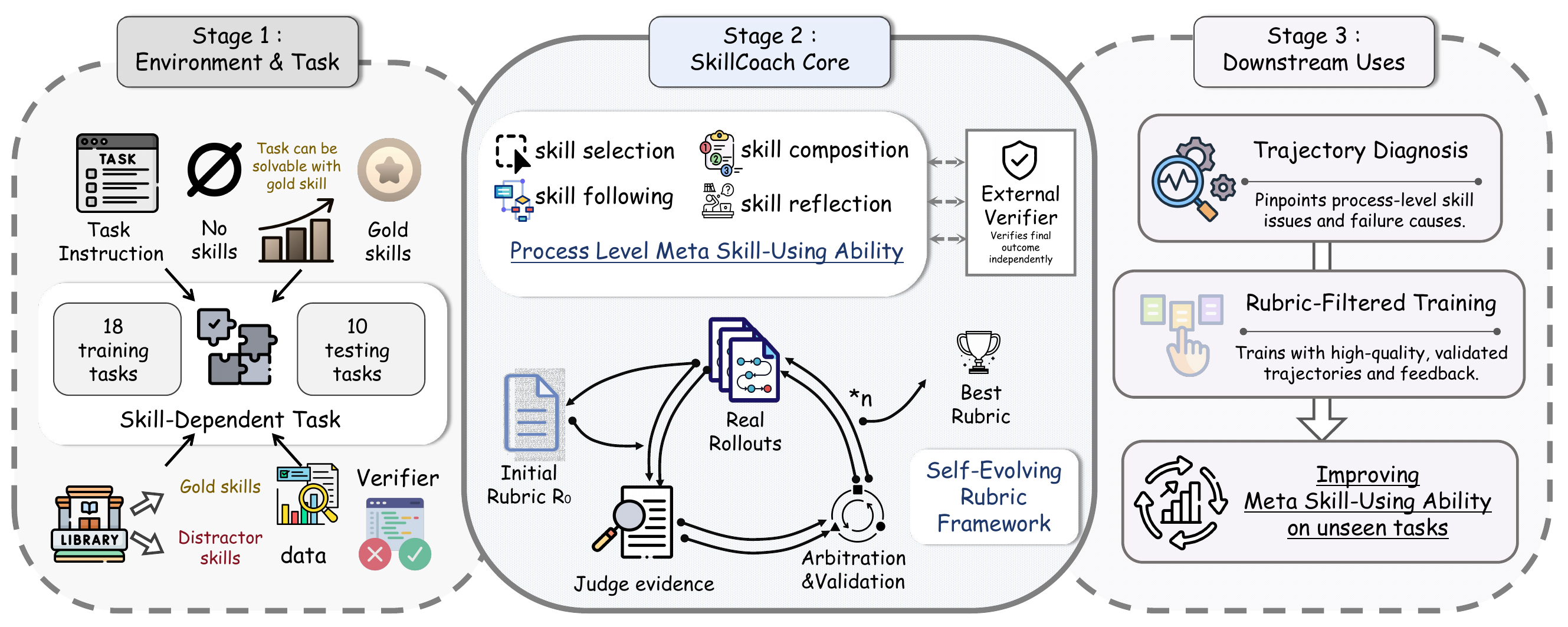}
  \caption{Overall framework of \textsc{SkillCoach}.}
  \label{fig:framework}

\end{figure*}

\section{Method}
\label{sec:method}

\subsection{Overall Framework}

Figure~\ref{fig:framework} shows \textsc{SkillCoach} addresses a deployment problem in enterprise skill repositories: given a skill-dependent task, an executable environment and a library with gold and distractor skills, it evaluates observable rollouts along four process dimensions---skill selection, skill following, skill composition  and skill-grounded reflection---while keeping the external verifier as a separate outcome signal. It then evolves task-specific trajectory-level rubrics from real rollouts through evidence-grounded judging, targeted arbitration patches, and validation-gated acceptance, and uses the best evolved rubric for each task to support trajectory diagnosis and rubric-filtered training.

\subsection{Skill-Dependent Tasks, Skills, and Trajectories}
\label{sec:task-definition}

We study skill-dependent tasks, which are difficult to solve reliably without the relevant gold skill but become substantially easier once the skill is available. This setting reflects enterprise deployments, where the key challenge is not merely whether useful skills exist, but whether agents can use them correctly.

To ensure that our tasks evaluate skill use rather than memorization or accidental verifier passing, we apply a skill-dependency filtering protocol. For each task, we estimate the no-skill success rate $p_{\mathrm{no}}$ and the gold-skill success rate $p_{\mathrm{gold}}$ across five strong agent backends, and compute the key-step coverage $c_{\mathrm{key}}$ with semi-automatic LLM-assisted annotation over successful gold-skill trajectories. A task is accepted only if
\begin{equation}
p_{\mathrm{no}}\leq \alpha,\quad
p_{\mathrm{gold}}-p_{\mathrm{no}}\geq \beta,\quad
c_{\mathrm{key}}\geq \gamma,
\end{equation}
where we set $\alpha=0.30$, $\beta=0.40$, and $\gamma=0.70$. As shown in Table~\ref{tab:skill_dependency_stats}, the selected tasks exhibit much larger gold-skill gains than the original SkillsBench pool~\citep{skillbench2026}, indicating that these tasks more directly test whether agents can use skills effectively.

We split these tasks into training and test sets at the task-family level, so that evaluation measures generalization to unseen skill-dependent workflows rather than memorization of task-specific trajectories. The full task inventory is provided in Appendix~\ref{app:task-inventory}. To approximate industrial skill repositories, we introduce distractor skills. In real deployments, agents often operate over a shared, non-task-isolated skill pool: a user query may surface multiple plausible skills, of which only some are applicable to the current task. In our default setting, each task is paired with two unrelated skills and three semantically similar but inapplicable distractors, as detailed in Appendix~\ref{app:task-inventory}. We further analyze more diverse distractors and their effects in Section~\ref{sec:distractor-boundary}.

Formally, each accepted task is represented as
\begin{equation}
\begin{aligned}
\tau_t &= (x_t,y_t,\mathcal{E}_t,\mathcal{V}_t,\mathcal{L}_t),\\
\mathcal{L}_t &= {\mathcal{G}}_t\cup\mathcal{D}_t.
\end{aligned}
\end{equation}
Here, $x_t$ is the task instruction, $y_t$ the target output or final state, $\mathcal{E}_t$ the executable environment, $\mathcal{V}_t$ the external verifier, ${\mathcal{G}}_t$ the exposed gold skills, and $\mathcal{D}_t$ the distractor skills. Each skill is centered on a \texttt{SKILL.md} file with optional scripts or references; its metadata specifies when it should be invoked, while its content defines workflow steps and validation requirements. The agent $\pi_{\theta}$ interacts with the environment and produces an \textbf{observable interaction trajectory} $z_t$:
\begin{equation}
z_t=(a_1,o_1,\ldots,a_T,o_T)
\sim
\pi_{\theta}(\cdot\mid x_t,\mathcal{E}_t,\mathcal{L}_t),
\end{equation}
where actions and observations include skill-document reads, skill invocations,
tool calls, file operations, script executions, intermediate artifacts, and final
outputs. The verifier is applied only after trajectory completion, and its result
is recorded as a post-hoc evaluation signal; $\mathcal{V}_t$ is not available to
the agent during interaction.

\begin{table}[t]
\centering

\small
\setlength{\tabcolsep}{4.5pt}
\renewcommand{\arraystretch}{1.08}
\begin{tabular}{@{}lccc@{}}
\toprule
\textbf{Task Set} & \textbf{$p_{\mathrm{no}}$} & \textbf{$p_{\mathrm{gold}}$} & \textbf{Gain (pp)} \\
\midrule
SkillsBench tasks & 33.8\% & 47.2\% & +13.4 \\
Selected training tasks & 19.6\% & 74.8\% & +55.2 \\
Selected test tasks & 16.8\% & 67.6\% & +50.8 \\
\bottomrule
\end{tabular}
\caption{Skill-dependency statistics. Selected tasks show lower no-skill success and substantially larger gains from gold skills.}

\label{tab:skill_dependency_stats}
\end{table}

\subsection{Agentic Meta Skill-Using Ability}
\label{Agentic Skill-Use Meta Ability}

We define agentic skill-use as the trajectory-level meta-ability to select, follow, compose, and reflect on skills under a given task and candidate skill library. This ability is distinct from final task success: a trajectory may pass the verifier through trial-and-error without properly using the intended skill. We therefore evaluate both process quality and the external outcome signal.

\paragraph{Skill selection.}
Skill selection evaluates whether the agent invokes the appropriate gold skills while avoiding distractors. Let $\widehat{\mathcal{S}}(z_t)$ denote the set of skills selected by the agent, extracted from visible evidence such as reading a skill document or explicitly invoking a skill. We define
\begin{equation}
s_{\mathrm{sel}}(z_t)=
\begin{cases}
\dfrac{2|\widehat{\mathcal{S}}(z_t)\cap \mathcal{G}_t|}
{|\widehat{\mathcal{S}}(z_t)|+|\mathcal{G}_t|+\epsilon},
& |\mathcal{G}_t|>0,\\[1.0em]
\mathbb{I}[\widehat{\mathcal{S}}(z_t)=\emptyset],
& |\mathcal{G}_t|=0.
\end{cases}
\end{equation}
The first case is a set-level F1 score that penalizes both missing required skills and selecting distractors. The second case captures negative abstention: when no applicable skill exists, the agent should not force the use of a similar but irrelevant skill.

\paragraph{Skill following.}
Skill following evaluates whether the agent follows the key procedures specified by the gold skills and reference solution. Let $\mathcal{K}_t$ be the set of rubric-defined key steps. For each step $k\in\mathcal{K}_t$, $c_k(z_t)\in\{0,0.5,1\}$ denotes its observed completion level, and $m_k(z_t)\in\{0,1\}$ indicates whether the completion is supported by visible trajectory evidence. The following score is
\begin{equation}
s_{\mathrm{fol}}(z_t)=
\frac{
\sum_{k\in\mathcal{K}_t} w_k\, c_k(z_t)\, m_k(z_t)
}{
\sum_{k\in\mathcal{K}_t} w_k
}.
\end{equation}
Here, $w_k$ is the importance weight of step $k$. The evidence multiplier $m_k$ prevents a trajectory from receiving credit for merely claiming that a step was completed. Missing key steps, unsupported claims, and skipped script executions all reduce the score.

\paragraph{Skill composition.}
Skill composition evaluates whether the agent coordinates multiple skills or dependent subprocesses in a valid workflow. Let $\mathcal{P}_t$ denote the set of precedence dependencies for task $t$, where $(u,v)\in\mathcal{P}_t$ means that step or skill $u$ should be completed before $v$. We score composition as
\begin{equation}
s_{\mathrm{comp}}(z_t)=
\frac{
\sum_{(u,v)\in\mathcal{P}_t}\beta_{uv}\,q_{uv}(z_t)
}{
\sum_{(u,v)\in\mathcal{P}_t}\beta_{uv}
}.
\end{equation}
Here, $q_{uv}(z_t)\in[0,1]$ indicates whether the trajectory completes $u$ before $v$ and correctly transfers the required intermediate artifact, state or output. For single-skill tasks or instances without explicit dependencies, this dimension is marked as not applicable and excluded from aggregation.

\paragraph{Skill-grounded result reflection.}
This dimension evaluates whether the agent performs explicit checks before final submission. Reflection is not equivalent to verifier success. Let $\mathcal{H}_t$ denote the set of expected checks, such as validating output files, schemas, formats or task-specific constraints. We define
\begin{equation}
s_{\mathrm{ref}}(z_t)=
\frac{
\sum_{c\in\mathcal{H}_t}\rho_c\,r_c(z_t)
}{
\sum_{c\in\mathcal{H}_t}\rho_c
},
\quad
r_c(z_t)\in\{0,0.5,1\}.
\end{equation}
Here, $r_c(z_t)$ measures the observed quality of check $c$. Together, these four dimensions define the process-level view of agentic skill-use. The scalar aggregation used for filtering and its separation from the external verifier are defined in Appendix~\ref{app:process-score}.

\subsection{Self-Evolving Trajectory-Level Rubrics}

\begin{figure}[t]
  \centering
\includegraphics[width=0.65\columnwidth]{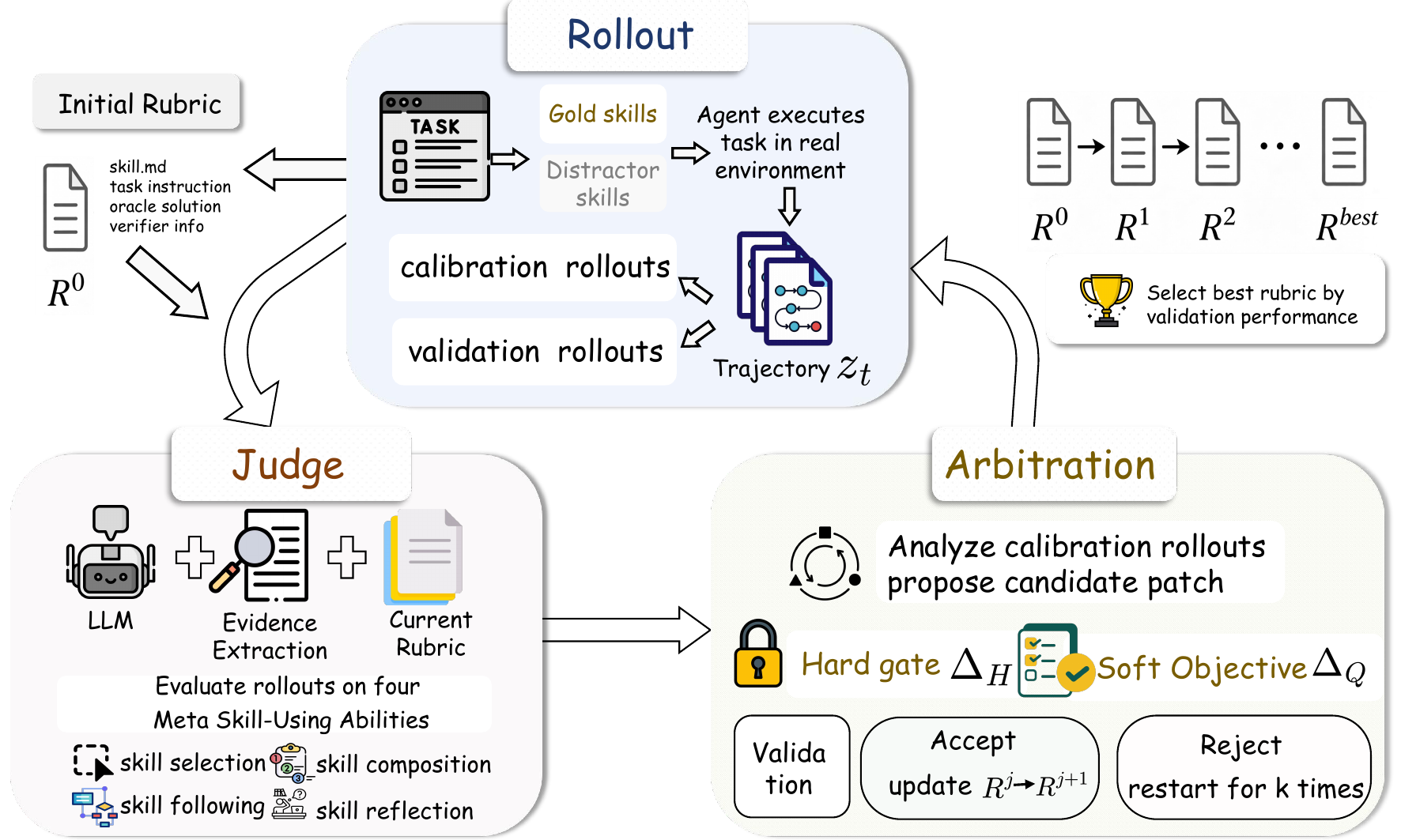}
  \caption{The Self-Evolving Rubric Framework.}
  \label{fig:Rubric framework}
\end{figure}
In enterprises, skills encode SOPs, domain rules, scripts, and validations. As repositories grow, task-specific rubrics become expensive to maintain and fragile when skills overlap. As shown in Figure~\ref{fig:Rubric framework}, \textsc{SkillCoach} treats each rubric as a versioned external evaluator and evolves it through three stages. First, the rollout stage collects real agent trajectories in the distractor-augmented library $\mathcal{L}_t=\mathcal{G}_t\cup\mathcal{D}_t$. Second, the judge stage applies the current rubric to trajectory evidence and scores the process dimensions. Third, the arbitration stage proposes patches and accepts them only through validation-gated checks. For each accepted task $\tau_t$, we maintain $R_t^j=\{R_{t,d}^j\}_{d\in\mathcal{M}_t}$, where $\mathcal{M}_t\subseteq\{\mathrm{sel},\mathrm{fol},\mathrm{comp},\mathrm{ref}\}$ denotes skill selection, following, composition, and skill-grounded reflection. The verifier $\mathcal{V}_t$ remains separate from the rubric.

\paragraph{Initial construction.}
We build $R_t^0$ from the task instruction, candidate skills, gold and distractor skills, verifier information, and representative rollouts. It is grounded in observable key steps $\mathcal{K}_t=\{k_1,\ldots,k_m\}$, where each $k_i$ maps to visible events such as file reads, script calls, artifacts, final writes, or checks. The rubric defines scoring rules, evidence requirements, and negative cases, each with a stable identifier for local patching.

\paragraph{Trajectory judging.}
The rollout agent $\pi_\theta$ executes in the actual environment and produces a trajectory with a verifier result:
\begin{equation}
\begin{aligned}
z_{t,i}&\sim \pi_\theta(\cdot\mid x_t,\mathcal{E}_t,\mathcal{L}_t),\\
v_{t,i}&=\mathcal{V}_t(\mathrm{final}(z_{t,i})),\\
\mathbf{s}_{t,i}^{j}&=J_\phi(R_t^j,z_{t,i},E(z_{t,i})).
\end{aligned}
\end{equation}
Here $E(z)=\mathrm{ExtractEvidence}(z)$ extracts skill reads, gold/distractor signals, tool calls, file edits, script executions, artifacts, and validation attempts. Real rollouts expose defects such as wrong-skill selection, skipped scripts, wrong skill order, and verifier-passing trial-and-error. Skill selection is rule-dominant and requires visible skill evidence. For following, composition, and reflection, positive judgments must cite event evidence; the verifier is only an external outcome signal.

\paragraph{Validation-gated evolution.}
At each round, rollouts are split into calibration $\mathcal{C}_t$ and held-out validation $\mathcal{U}_t$. A separate arbitration model proposes a local patch:
\begin{equation}
\widetilde{R}_t^{j+1}=R_t^j\oplus A_\psi(R_t^j,\mathcal{C}_t).
\end{equation}
The patch may add evidence requirements, negative cases, scoring constraints, ordering dependencies, or optional-step clarifications; it cannot inspect validation rollouts, accept itself, bypass the verifier, or delete critical key steps.

The candidate is evaluated on $\mathcal{U}_t$. The hard gate $H$ rejects destructive changes, such as ignoring distractors or marking unsupported selection or key-step completion as correct. The soft objective $Q$ measures evidence coverage, evidence quality, reflection grounding, process-verifier consistency, and compactness. Let $\Delta H$ and $\Delta Q$ compare the candidate against $R_t^j$. We accept $\widetilde{R}_t^{j+1}$ only if $\Delta H\ge0$, $\Delta Q>\epsilon$, $\Delta_{\mathrm{mat}}\neq\varnothing$, and no structural violation is detected, where $\Delta_{\mathrm{mat}}$ requires material improvement beyond increased judge confidence. 
\begin{equation}
\begin{gathered}
\mathcal{A}_t=\{R_t^0\}\cup\{R:\operatorname{Accept}(R)=1\},\\
R_t^{\mathrm{best}}=\arg\max_{R\in\mathcal{A}_t}Q(R,\mathcal{U}_t).
\end{gathered}
\end{equation}
The final rubric is applied for trajectory diagnosis, filtering training data, and assigning rewards. Implementation details and prompts are in Appendix~\ref{app:prompt-templates}.

\subsection{Training with Rubric-Filtered Trajectories}

We use the selected evolved rubric $R^{\mathrm{best}}$ as an offline process-quality filter for supervised skill-use training. For each task, we collect rollouts with the distractor-augmented library $\mathcal{L}_t$ and retain a trajectory only when it attains a rubric-based meta score of at least $0.95$ and passes the external verifier. The resulting set is denoted as $\mathcal{D}_{\mathrm{sft}}$. Training details are provided in Appendix~\ref{app:sft-training}.

\section{Experiments}
\label{sec:experiments}

We evaluate \textsc{SkillCoach} through four questions that reflect the full deployment path of skill-using agents. First, we test whether the evolved rubrics are faithful and usable trajectory-level evaluation instruments. Second, we examine whether our four-dimensional evaluation of meta skill-using ability reveals failures that final task accuracy alone cannot diagnose under distractor-augmented skill-library conditions. Third, we evaluate whether trajectories filtered by self-evolved rubrics improve trainable agentic skill-use ability. Finally, we stress-test skill selection as the candidate library scales from gold-only settings to large shared skill repositories. Unless otherwise specified, we evaluate on the training and test skill-dependent tasks defined in Section~\ref{sec:task-definition}, with task inventory and setup details reported in Appendix~\ref{app:task-inventory}.
\begin{table}[t]
\centering
\small
\setlength{\tabcolsep}{7pt}
\renewcommand{\arraystretch}{1.08}
\begin{tabular}{lccc}
\hline
\textbf{Evaluation Metric}
& \textbf{$R^0$}
& \textbf{$R^{\mathrm{best}}$}
& \textbf{Change} \tabularnewline
\hline
gold-keypoint coverage
& 71.56
& 83.70
& +12.14 pp \tabularnewline
Usability score
& 81.53
& 94.33
& +12.80 points \tabularnewline
Hallucination rate
& 2.00
& 0.00
& -2.00 pp \tabularnewline
Filtering consistency
& 82.00
& 96.00
& +14.00 pp \tabularnewline
\hline
\end{tabular}
\caption{
Human-gold validation of rubrics before and after self-evolution.
}
\label{tab:rubric-validation}
\end{table}

\subsection{Quality Validation of Self-Evolving Rubrics}
\label{sec:rubric-validation}

We first evaluate whether the self-evolving rubric itself is a faithful trajectory-level evaluator. This differs from task evaluation: a verifier score only measures whether the final artifact passes, whereas a rubric specifies what process evidence a trajectory judge should inspect, which skills should be used, which steps should be followed, and which failure modes should be penalized. We therefore exclude verifier pass rates and final task outcomes from this validation.

We compare the initial rubric $R^0$ and the evolved rubric $R^{\mathrm{best}}$ on 50 paired test instances from 10 held-out task families. For each task, the two rubrics are evaluated under the same task package and the same human-gold process reference. The reference is constructed by human annotators from oracle materials, covering gold skills, distractor boundaries, key steps, skill-composition order, and reflection criteria. To avoid overlap between rubric generation and rubric evaluation, we use Gemini 3.1 Pro with temperature 0 as an independent rubric-quality judge. The judge receives the task description, gold skills, distractor skills, human-gold process evidence, ordering constraints, reflection criteria, and one candidate rubric, and is instructed to ignore verifier pass/fail signals when judging rubric quality. All values in Table~\ref{tab:rubric-validation} are task-level macro-averages over the 50 paired instances.

We report four rubric-quality metrics. \textbf{Gold-keypoint coverage} measures the fraction of human-gold keypoints that the rubric can explicitly evaluate. A keypoint is counted as covered only when the rubric contains task-specific criteria or evidence requirements that allow a trajectory judge to recognize whether that keypoint was completed. \textbf{Usability score} measures whether the rubric can be directly used by a trajectory judge; it averages completeness, logical consistency, and judge usability on a 0--100 scale. \textbf{Hallucination rate} measures the fraction of non-verifier rubric requirements that introduce unsupported task constraints, such as wrong skill requirements, wrong file paths, wrong output formats, conflicting ordering constraints, or irrelevant generic requirements. \textbf{Trajectory-filtering consistency} measures whether the rubric supports the same accept/reject decision as the human-gold trajectory judgment, and is not a verifier score.

Table~\ref{tab:rubric-validation} shows that $R^{\mathrm{best}}$ improves all four metrics. Gold-keypoint coverage increases from 71.56 to 83.70, indicating that self-evolution makes rubrics more complete with respect to human-gold process evidence. Usability improves from 81.53 to 94.33, suggesting that evolved rubrics provide clearer criteria, evidence requirements, and negative cases for LLM-based trajectory judging. Hallucination rate decreases from 2.00 to 0.00, showing that the gain in coverage does not come from adding unsupported requirements. Trajectory-filtering consistency improves from 82.00 to 96.00, indicating stronger alignment with gold trajectory-level judgments. This validation should be interpreted as a human-gold, judge-assisted audit rather than a new human annotation study. The paired design, task-level macro-averaging, separation between rubric generation and rubric evaluation models, and exclusion of verifier outcomes are intended to reduce the main validity threats of this analysis. Overall, these results suggest that rubric self-evolution improves the rubric as an evaluation artifact, not merely as a proxy for final verifier success.

\subsection{Overall Agentic Skill-Use Performance under Skill Libraries}
\label{sec:overall-performance}

We next evaluate whether agents can use skills under different skill-library conditions using human-annotated gold rubrics that cover our meta-ability dimensions. In the Gold + Distractors setting, each task keeps its gold skills while adding two unrelated skills and three semantically similar but inapplicable distractors. Table~\ref{tab:overall-performance} shows that skills improve task success, but do not guarantee reliable skill use.

Moving from No Skills to Gold Skills substantially raises final accuracy for strong agents, confirming that the selected tasks are skill-dependent rather than merely testing generic problem solving. For example, Opus 4.7 improves from 18.0 to 88.0 final accuracy, and GPT-5.5 improves from 24.0 to 80.0. This indicates that reusable procedural knowledge is necessary for solving these tasks reliably. However, adding distractors reduces performance for most agents even though the gold skills remain available. This shows that the bottleneck is not skill availability alone, but reliable skill selection and execution under a realistic candidate space.

The four skill-use dimensions clarify the failure mode. Gemini 3.1 Pro maintains strong following performance under distractors, but its selection score drops from 98.0 to 78.0. Qwen3.5-9B shows an even sharper selection drop, from 92.0 to 44.0, accompanied by lower reflection and final accuracy. These patterns suggest that once an agent reaches the correct skill, it may still execute part of the procedure, but finding the correct skill becomes fragile when the library contains plausible alternatives. Final accuracy alone would hide this distinction: a lower verifier score does not reveal whether the failure comes from wrong skill choice, incomplete step following, weak composition, or missing checks. Conversely, a trajectory may preserve useful process behavior even when the final outcome fails.

These results support the need for joint outcome-process evaluation. In enterprise skill repositories, skill overlap, false activation, and near-miss skill selection are expected rather than exceptional. A final successful outcome may still be produced through ad hoc recovery after an incorrect skill invocation, while the underlying trajectory remains costly, brittle, or non-reusable. By separating final accuracy from skill selection, key-step following, composition order, and reflection, \textsc{SkillCoach} provides a more operational diagnosis of agentic skill-use failures under shared skill libraries.

\begin{table*}[t]
\centering
\small
\setlength{\tabcolsep}{6pt}
\renewcommand{\arraystretch}{1.08}
\definecolor{DistractorRow}{HTML}{F2F2F2}

\begin{tabularx}{\textwidth}{
@{}
>{\raggedright\arraybackslash}p{0.18\textwidth}
>{\raggedright\arraybackslash}p{0.15\textwidth}
>{\centering\arraybackslash}X
>{\centering\arraybackslash}X
>{\centering\arraybackslash}X
>{\centering\arraybackslash}X
>{\centering\arraybackslash}X
@{}
}
\toprule
\textbf{Model}
& \textbf{Skill Library}
& \textbf{Final}
& \textbf{Sel.}
& \textbf{Follow}
& \textbf{Order}
& \textbf{Reflect} \\
\midrule

\textbf{DeepSeek-V4-Flash}
& No Skills
& 10.0
& \multicolumn{4}{c}{--} \\
& Gold Skills
& 32.0 & 88.0 & 40.1 & 68.9 & 72.0 \\
\rowcolor{DistractorRow}
& Gold + Distr.
& 26.0 & 66.0 & 39.6 & 62.1 & 86.0 \\
\addlinespace[0.35em]

\textbf{Opus-4.7}
& No Skills
& 18.0
& \multicolumn{4}{c}{--} \\
& Gold Skills
& 88.0 & 100.0 & 86.5 & 100.0 & 90.0 \\
\rowcolor{DistractorRow}
& Gold + Distr.
& 80.0 & 90.0 & 77.2 & 89.7 & 90.0 \\
\addlinespace[0.35em]

\textbf{GPT-5.5}
& No Skills
& 24.0
& \multicolumn{4}{c}{--} \\
& Gold Skills
& 80.0 & 100.0 & 80.0 & 96.6 & 100.0 \\
\rowcolor{DistractorRow}
& Gold + Distr.
& 76.0 & 96.0 & 71.8 & 93.1 & 84.0 \\
\addlinespace[0.35em]

\textbf{Gemini-3.1-Pro}
& No Skills
& 16.0
& \multicolumn{4}{c}{--} \\
& Gold Skills
& 72.0 & 98.0 & 72.6 & 82.8 & 90.0 \\
\rowcolor{DistractorRow}
& Gold + Distr.
& 70.0 & 78.0 & 82.7 & 89.7 & 70.0 \\
\addlinespace[0.35em]

\textbf{Kimi-K2.6}
& No Skills
& 16.0
& \multicolumn{4}{c}{--} \\
& Gold Skills
& 66.0 & 92.0 & 84.2 & 93.1 & 88.0 \\
\rowcolor{DistractorRow}
& Gold + Distr.
& 54.0 & 86.0 & 72.7 & 86.2 & 86.0 \\
\addlinespace[0.35em]

\textbf{Qwen3.5-4B}
& No Skills
& 0.0
& \multicolumn{4}{c}{--} \\
& Gold Skills
& 8.0 & 80.0 & 39.5 & 44.8 & 64.0 \\
\rowcolor{DistractorRow}
& Gold + Distr.
& 8.0 & 44.0 & 36.7 & 41.4 & 38.0 \\
\addlinespace[0.35em]

\textbf{Qwen3.5-9B}
& No Skills
& 2.0
& \multicolumn{4}{c}{--} \\
& Gold Skills
& 18.0 & 92.0 & 47.2 & 48.3 & 76.0 \\
\rowcolor{DistractorRow}
& Gold + Distr.
& 14.0 & 44.0 & 44.5 & 34.5 & 34.0 \\

\bottomrule
\end{tabularx}

\caption{
Overall performance under three skill-library settings. All values are percentages.
}
\label{tab:overall-performance}
\end{table*}

\begin{table}[t]
\centering
\footnotesize
\setlength{\tabcolsep}{6pt}
\renewcommand{\arraystretch}{1.08}
\definecolor{GroupGray}{HTML}{F6F6F6}
\definecolor{BestGray}{HTML}{EAEAEA}

\begin{tabular}{@{}lcc@{}}
\toprule
\textbf{Training / Model Setting}
& \textbf{4B}
& \textbf{9B} \\
\midrule

\rowcolor{GroupGray}
\multicolumn{3}{@{}l}{\textit{Baselines}} \\
Base Qwen3.5
& 8.0
& 14.0 \\
Outcome-only SFT
& 6.0
& 18.0 \\
DeepSeek-V4-Flash \textit{(284B ref.)}
& \multicolumn{2}{c@{}}{26.0} \\
\midrule

\rowcolor{GroupGray}
\multicolumn{3}{@{}l}{\textit{Rubric-filtered SFT}} \\
Full SFT with $R^0$
& 16.0
& 28.0 \\
\rowcolor{BestGray}
Full SFT with $R^{\mathrm{best}}$
& \textbf{24.0}
& \textbf{32.0} \\
\midrule

\rowcolor{GroupGray}
\multicolumn{3}{@{}l}{\textit{Meta-ability filtering ablations}} \\
w/o composition order
& 18.0
& 26.0 \\
w/o reflection
& 24.0
& 28.0 \\
w/o key-step following
& 10.0
& 16.0 \\

\bottomrule
\end{tabular}

\caption{
SFT ablation under the Gold + Distractors setting. Results are final accuracy (\%).
}
\label{tab:sft-ablation}
\end{table}

\subsection{Rubric-Filtered Training and Meta-Ability Ablations}
\label{sec:sft-ablation}

We then test whether evolved rubrics can improve agentic skill-use ability by selecting better supervised trajectories. Table~\ref{tab:sft-ablation} reports final accuracy under the Gold + Distractors setting. The DeepSeek V4 Flash row is included as a large reference agent, not as a trained variant. The baseline Qwen3.5 agents perform poorly under this setting, indicating that smaller open models struggle with realistic skill-use conditions.

Rubric-filtered SFT improves both model scales. Filtering trajectories with the initial rubric $R^0$ increases Qwen3.5-4B from 8.0 to 16.0 and Qwen3.5-9B from 14.0 to 28.0. Using the evolved rubric $R^{\mathrm{best}}$ further improves performance to 24.0 and 32.0, respectively. This supports the value of rubric evolution: a better evaluator selects better demonstrations. In contrast, outcome-only SFT is weaker. It reduces the 4B result from 8.0 to 6.0 and provides only a modest gain for 9B, from 14.0 to 18.0. This confirms our motivation that verifier-passing trajectories are not automatically reusable skill-use examples. A trajectory can pass the final checker while selecting a distractor, skipping the intended SOP, or reaching the answer through non-reusable trial and error. Rubric-filtered SFT instead keeps trajectories that satisfy both outcome and process constraints.

The ablations further show which process criteria are most useful for filtering. Skill selection is treated as a prerequisite for trajectory filtering rather than a removable scoring term, so it is not included as an ablation variant. Among the tested criteria, removing key-step following causes the largest drop, reducing performance to 10.0 for 4B and 16.0 for 9B. Removing composition order also reduces performance at both scales, while removing reflection has a smaller but visible effect, especially for 9B. These results show that the tested process criteria provide useful supervision signals, with faithful execution of skill procedures contributing the most. The main conclusion is not that one trained model is stronger than another; rather, the experiment demonstrates that agentic skill-use is trainable, and that evolved trajectory-level rubrics can turn process diagnosis into effective data selection for improving skill-use behavior.

\begin{figure}[t]
\centering
\includegraphics[width=\linewidth]{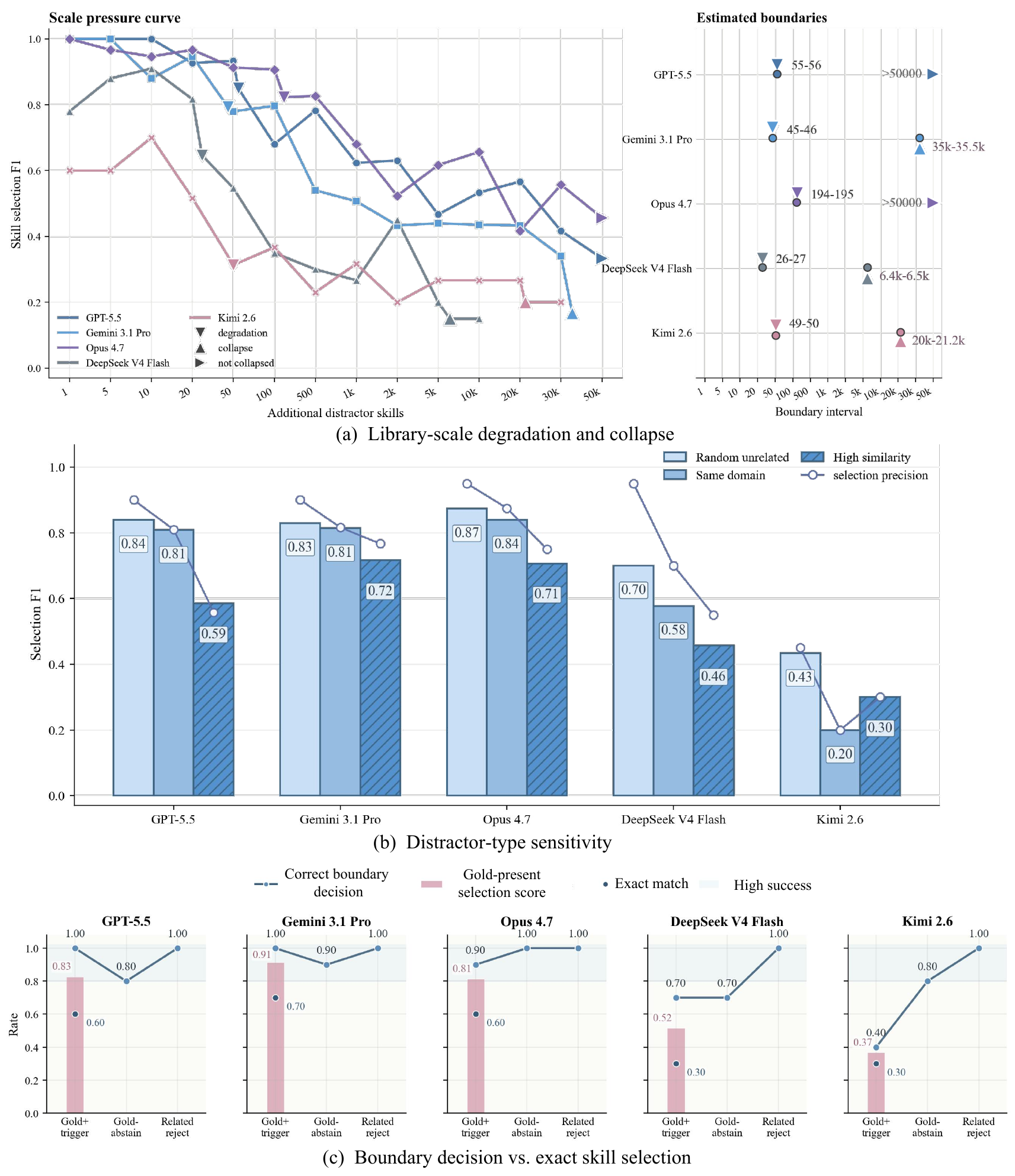}
\caption{
Distractor-boundary analysis under growing and semantically overlapping skill libraries.}
\label{fig:distractor-boundary}
\end{figure}

\subsection{Distractor-Boundary Analysis: Scaling Skill Selection in Large Skill Libraries}
\label{sec:distractor-boundary}

We further study the reliability boundaries of skill selection in large shared skill libraries. Starting from a clean gold-only candidate set, we progressively expand the library with real distractor skills and measure two breakpoints: when selection quality first degrades and when models almost fail to select the gold skills. This setting approximates production skill repositories, where skills are shared across scenarios and a user query may surface multiple plausible candidates, only some of which are applicable to the current task. Figure~\ref{fig:distractor-boundary} reports a distractor-boundary analysis that deliberately isolates the skill-selection dimension rather than running the full task-container execution. Models inspect candidate skill documents through a constrained browser-style interface with \texttt{list}, \texttt{search}, \texttt{read}, and \texttt{final} actions, and finally output selected skill identifiers or abstain from skill use. This design removes confounding factors from tool execution, environment failures, and verifier noise, allowing us to directly probe whether a model can identify applicable gold skills in a large and overlapping candidate space.

This setting reflects a common industrial deployment challenge. Many controlled skill-use evaluations~\cite{skillbench2026,skilllearnbench2026} place only the gold skills in the task environment, thereby primarily testing whether an agent can invoke and follow a provided skill. In real enterprise settings, however, user queries usually express intentions rather than explicitly naming the skill to invoke. The agent must infer whether skill use is needed, route the query to the appropriate skill in a shared business library when applicable, distinguish it from semantically similar alternatives, and abstain when no skill applies. As teams accumulate skills for reporting, data cleaning, compliance checking, dependency auditing, document processing, and other recurring workflows, the shared skill pool often contains skills with similar trigger words, related business objects, or partially overlapping procedures. The central question therefore shifts from whether a useful skill exists to whether the agent can identify the right skill among many plausible alternatives. We therefore use distractor skills as a stress test that bridges gold-only evaluation and production-scale skill libraries.

We define two boundary notions. The \textbf{degradation boundary} is the first library size where selection F1 drops by at least 0.10 from the no-distractor baseline and does not recover at larger tested sizes. The \textbf{collapse boundary} is the first stable scale where the model almost fails to recover gold skills. We operationalize collapse by any of the following criteria: complete failure rate or no-gold-selected rate reaches 0.80, or selection F1 is no higher than 0.10 while exact match is no higher than 0.05. Here, complete failure means that gold skills are present but the selected set has no overlap with the gold set. Thus, collapse does not require the average F1 to be exactly zero: a model can still obtain a nonzero mean F1 if a small fraction of samples partially recover gold skills, while most samples already fail to recover any gold skill. The main scale experiment uses real skill documents rather than synthetic skill generation. The official Skill-Usage pool~\cite{liu2026agenticskillsworkwild} yields 35,554 valid \texttt{SKILL.md} files after materialization and duplicate handling; we further extend it with additional SkillHub-derived skills to form the 50k stress-test library.

Figure~\ref{fig:distractor-boundary}(a) shows that degradation and collapse are distinct phenomena. We first estimate each model's selection ability without distractors and then gradually increase the number of real distractor skills. Even strong models enter a lower-reliability region after relatively small candidate expansions. Gemini 3.1 Pro and GPT-5.5 reach the degradation boundary around 45--46 and 55--56 distractors, respectively, while Opus 4.7 degrades later, around 194--195 distractors. DeepSeek V4 Flash degrades earlier, around 26--27 distractors. Kimi K2.6 starts from a weaker no-distractor baseline, so later drops should not be attributed solely to library scale, but its curve still indicates limited selection stability.

Collapse occurs much later and is primarily driven by complete failures rather than the F1-only criterion. DeepSeek V4 Flash collapses around 6.4k--6.5k distractors, Kimi K2.6 around 20k--21.25k, and Gemini 3.1 Pro around 35k--35.5k. At the corresponding collapse boundary points, their mean selection F1 remains nonzero: DeepSeek V4 Flash has F1 0.15 and exact match 0.10 at 6416 distractors, Gemini 3.1 Pro has F1 0.17 and exact match 0.10 at 35,500 distractors, and Kimi K2.6 has F1 0.20 and exact match 0.20 at 21,250 distractors. However, in all three cases, both no-gold-selected rate and complete failure rate reach 0.80, meaning that 80\% of samples already select no gold skill and have no overlap with the gold set. In contrast, GPT-5.5 and Opus 4.7 do not meet the collapse criteria at the official-pool scale and remain outside the collapse regime even at the extended 50k stress-test scale. At 50k distractors, GPT-5.5 still has F1 0.33, exact match 0.20, no-gold-selected rate 0.60, and complete failure rate 0.60, while Opus 4.7 has F1 0.46, exact match 0.30, no-gold-selected rate 0.40, and complete failure rate 0.40. This separation is important for deployment: degradation suggests that retrieval, ranking, or candidate filtering should be improved, while collapse indicates that the library has exceeded the range in which the model can make reliable skill selections.

Figure~\ref{fig:distractor-boundary}(b) shows that the semantic type of distractors matters as much as their number. With 50 distractor skills fixed, we compare random unrelated, same-domain, and high-similarity distractors. Random unrelated skills are relatively less harmful, same-domain skills introduce moderate confusion, and high-similarity skills are the most damaging. For example, GPT-5.5 drops from 0.84 selection F1 under random unrelated distractors to 0.59 under high-similarity distractors; Opus 4.7 drops from 0.87 to 0.71; and DeepSeek V4 Flash drops from 0.70 to 0.46. This matches a common production risk: the most dangerous candidates are not obviously irrelevant tools, but plausible near-neighbor skills, such as similar reporting workflows across departments, different versions of a compliance checker, or automation scripts whose trigger conditions differ only by a small business constraint. These near-neighbor skills can induce over-selection, wrong selection, or partially correct but imprecise selection.

Figure~\ref{fig:distractor-boundary}(c) further separates boundary control from exact skill selection. We decompose selection into positive triggering, where gold skills are present and should be used, and negative abstention, where the request has no applicable gold skill or should not trigger skill use despite related distractors. Most models achieve high abstention rates on non-trigger requests with related distractors, suggesting that they do not simply invoke skills whenever a similar candidate appears. However, the gold-present case still reveals substantial differences in exact skill selection. Gemini 3.1 Pro obtains the strongest gold-present selection F1 in this setting, while GPT-5.5 and Opus 4.7 remain competitive. DeepSeek V4 Flash and Kimi K2.6 show weaker triggering and exact selection. This indicates that boundary control is not only about avoiding unnecessary skill use; it also requires selecting the complete and precise gold skill set when skill use is warranted. Measuring only whether an agent triggers any skill would therefore overestimate capability, because the agent may invoke a skill while missing required gold skills or adding irrelevant ones.

Taken together, Figure~\ref{fig:distractor-boundary} reveals a failure mode that gold-only skill evaluations largely miss. Agentic skill use is not merely the ability to choose a skill from a clean candidate set; it is the ability to make reliable decisions as the skill library grows, becomes semantically overlapping, and contains ambiguous triggering conditions. Overall, stronger models are not immune to distractors, but they tend to degrade later, avoid collapse at larger scales, and maintain better boundary behavior. This analysis explains why \textsc{SkillCoach} evaluates skill selection explicitly at the trajectory level rather than relying only on final task success. In industrial agent systems, a final successful outcome can hide an incorrect skill invocation followed by ad hoc self-repair, while the incorrect invocation itself may still introduce cost, latency, compliance risk, and non-reusable execution patterns. Distractor-boundary analysis therefore provides a production-oriented diagnostic: it asks not only whether a model can use skills under ideal conditions, but how far that ability remains reliable when the skill library becomes large, overlapping, and noisy.

\section{Conclusion}

We introduced \textsc{SkillCoach}, a self-evolving rubric framework for evaluating and improving agentic skill-use in realistic skill repositories. By separating trajectory-level process quality from external verifier success, \textsc{SkillCoach} diagnoses whether agents select, follow, compose, and reflect on skills rather than merely pass tasks by trial and error. Across skill-dependent tasks with distractor-augmented libraries, evolved rubrics become more faithful evaluators, reveal failures hidden by final accuracy, and select better demonstrations for SFT than outcome-only filtering. These results suggest that reliable skill-use is both measurable and trainable through skill-grounded process supervision.

\section{Limitations}

We identify two main limitations. First, our experiments are conducted on a selected set of skill-dependent tasks from existing skill benchmarks. Although these tasks cover diverse enterprise-style domains and are split at the task-family level to test generalization to unseen workflows, the scale remains smaller than large production skill repositories with continuously growing tasks, tools, and operational procedures. Second, our training study focuses on offline supervised fine-tuning with rubric-filtered trajectories. We do not report on-policy reinforcement learning or long-term deployment feedback in this public version. The process scores produced by \textsc{SkillCoach} could naturally serve as reward signals for RL, which we leave to future work.

\bibliographystyle{unsrtnat}
\bibliography{neurips_2026}


\appendix

\renewcommand{\catrule}{\specialrule{0.28pt}{2pt}{2pt}}

\begin{table*}[t]
\centering
\small
\setlength{\tabcolsep}{5.2pt}
\renewcommand{\arraystretch}{1.08}

\begin{tabularx}{\textwidth}{
@{}
>{\raggedright\arraybackslash}p{0.21\textwidth}
>{\raggedright\arraybackslash}X
>{\centering\arraybackslash}p{0.06\textwidth}
>{\centering\arraybackslash}p{0.065\textwidth}
>{\centering\arraybackslash}p{0.06\textwidth}
@{}}
\toprule
\textbf{Category}
& \textbf{Task}
& \multicolumn{2}{c}{\textbf{Skill Library}}
& \textbf{Data} \\
\cmidrule(lr){3-4}
\cmidrule(l){5-5}
& & \textbf{Gold} & \textbf{Distr.} & \textbf{Inst.} \\
\midrule

\rowcolor{SkillSection}
\multicolumn{5}{@{}l}{\textbf{Training Tasks}} \\

\multirow{3}{=}{Software Engineering}
& software-dependency-audit & 3 & 5 & 3 \\
& fix-security-bug & 1 & 5 & 1 \\
& fix-erlang-ssh-cve & 6 & 5 & 1 \\
\catrule

\multirow{2}{=}{Productivity Tools}
& offer-letter-generator & 1 & 5 & 5 \\
& court-form-filling & 1 & 5 & 1 \\
\catrule

\multirow{3}{=}{Environmental \& Scientific Analysis}
& earthquake-plate-calculation & 1 & 5 & 1 \\
& flood-risk-analysis & 3 & 5 & 1 \\
& lake-warming-attribution & 4 & 5 & 1 \\
\catrule

\multirow{4}{=}{Business \& Financial Analytics}
& weighted-gdp-calc & 1 & 5 & 5 \\
& sales-pivot-analysis & 2 & 5 & 6 \\
& sec-financial-report & 2 & 5 & 5 \\
& invoice-fraud-detection & 3 & 5 & 4 \\
\catrule

\multirow{3}{=}{Scientific Computing}
& lab-unit-harmonization & 1 & 5 & 3 \\
& mars-clouds-clustering & 3 & 5 & 1 \\
& protein-expression-analysis & 1 & 5 & 5 \\
\catrule

\multirow{2}{=}{Optimization \& Operations}
& grid-dispatch-operator & 3 & 5 & 1 \\
& powerlifting-coef-calc & 3 & 5 & 5 \\
\catrule

Content \& Creative
& mario-coin-counting & 3 & 5 & 1 \\
\catrule

\rowcolor{SkillTotal}
\textbf{Total}
& \textbf{18 training tasks}
& \textbf{42}
& \textbf{90}
& \textbf{50} \\ \\
\midrule

\rowcolor{SkillSection}
\multicolumn{5}{@{}l}{\textbf{Test Tasks}} \\

\multirow{2}{=}{Software Engineering}
& nlp-paper-reproduction & 2 & 5 & 3 \\
& dependency-vulnerability-check & 3 & 5 & 5 \\
\catrule

Information Retrieval
& enterprise-information-search & 1 & 5 & 6 \\
\catrule

\multirow{3}{=}{Data \& Analytics}
& financial-analysis & 2 & 5 & 6 \\
& dbscan-parameter-tuning & 3 & 5 & 5 \\
& stock-data-visualization & 1 & 5 & 5 \\
\catrule

\multirow{3}{=}{Content \& Creative}
& anthropic-poster-design & 1 & 5 & 5 \\
& chinese-poem-generator & 1 & 5 & 5 \\
& video-object-counting & 3 & 5 & 5 \\
\catrule

Utilities \& Other
& temperature-simulation & 3 & 5 & 5 \\
\catrule

\rowcolor{SkillTotal}
\textbf{Total}
& \textbf{10 testing tasks}
& \textbf{20}
& \textbf{50}
& \textbf{50} \\

\bottomrule
\end{tabularx}

\caption{
\textsc{SkillCoach} task inventory. Gold denotes task-required skills, Distr. denotes distractor skills, and Inst. denotes task instances. Distractors include cross-task skills and semantically near but functionally inapplicable skills, simulating realistic industrial skill libraries.
}
\label{tab:skillcoach_tasks}
\end{table*}

\newpage
\appendix

\section{Task Inventory}
\label{app:task-inventory}

Table~\ref{tab:skillcoach_tasks} summarizes the training and test task inventory used in \textsc{SkillCoach}. We select all tasks from SkillsBench~\citep{skillbench2026} and SkillLearnBench~\citep{skilllearnbench2026} using the skill-dependency protocol described in Section~\ref{sec:task-definition}. The resulting inventory contains 18 training task families with 50 instances and 10 held-out test task families with 50 instances. We split tasks at the task-family level to evaluate generalization to unseen skill-dependent workflows rather than memorization of task-specific trajectories. The tasks cover diverse enterprise-style domains, including software engineering, productivity tools, scientific analysis, business analytics, optimization, and creative or multimodal tasks.

For rubric-quality validation and overall skill-use evaluation, we additionally construct human-gold rubrics for the test tasks. Two annotators build these references from the oracle materials and real rollouts, covering the four agentic skill-use dimensions: skill selection, key-step following, skill composition, and skill-grounded reflection. These human-gold rubrics are used only for evaluation, including the rubric-quality validation in Section~\ref{sec:rubric-validation}  and the overall agentic skill-use analysis in Section~\ref{sec:overall-performance}.

For each task, we report the task package composition, including the task instruction, executable environment, external verifier, oracle solution or reference workflow, task-required gold skills, injected distractor skills, and task instances. Each task uses a fixed distractor setting with two cross-task skills and three semantically near but functionally inapplicable skills. This setting simulates industrial skill repositories, where relevant and irrelevant skills may coexist and where overlapping descriptions or similar invoke conditions can lead to false activation or wrong skill selection.

To further study distractors at scale, we also expand the distractor library beyond this fixed setting. This simulates a deployment scenario in which an agent retrieves candidate skills from a large public or enterprise skill hub, where both the number and semantic similarity of available skills can vary substantially. We use this expanded setting to analyze the boundary of an agent's skill-selection robustness, as discussed in Section~\ref{sec:distractor-boundary}.

\section{Process-Score Aggregation and Verifier Signal}
\label{app:process-score}

Section~\ref{Agentic Skill-Use Meta Ability} defines the four skill-use dimensions separately because enterprise skill-use failures are often localized to a specific process dimension, such as wrong skill selection or missing key-step execution. For trajectory filtering, aggregate reporting, and later optimization, we additionally compute a scalar process score while preserving the external verifier as an independent outcome signal.

For the applicable process dimensions
$\mathcal{M}_t\subseteq\{\mathrm{sel},\mathrm{fol},\mathrm{comp},\mathrm{ref}\}$,
let $\hat{y}_t=\mathrm{final}(z_t)$ denote the final artifact or environment state. We compute
\begin{equation}
\begin{aligned}
S_{\mathrm{meta}}(z_t)
&=
\frac{\sum_{d\in\mathcal{M}_t}\lambda_d s_d(z_t)}
{\sum_{d\in\mathcal{M}_t}\lambda_d},\\
s_{\mathrm{ver}}(z_t)
&=
\mathcal{V}_t(\hat{y}_t)\in\{0,1,\bot\}.
\end{aligned}
\end{equation}
The default dimension weights are
$\lambda_{\mathrm{sel}}=0.40$,
$\lambda_{\mathrm{fol}}=0.30$,
$\lambda_{\mathrm{comp}}=0.20$, and
$\lambda_{\mathrm{ref}}=0.10$.
When a dimension is not applicable, such as composition in single-skill tasks, it is removed from $\mathcal{M}_t$ and the score is renormalized over the remaining applicable dimensions.

Here, $S_{\mathrm{meta}}(z_t)$ measures trajectory-level skill-use quality, while $s_{\mathrm{ver}}(z_t)$ records the external verifier result. The value $\bot$ denotes cases where the verifier cannot be executed or does not return a valid result. Each trajectory is summarized as
\begin{equation}
\mathbf{s}(z_t)
=
(s_{\mathrm{sel}},s_{\mathrm{fol}},s_{\mathrm{comp}},s_{\mathrm{ref}},s_{\mathrm{ver}}).
\end{equation}
This separation is important: a trajectory can pass the verifier while exhibiting poor skill use, and a failed trajectory may still contain useful process evidence. Therefore, \textsc{SkillCoach} uses $S_{\mathrm{meta}}$ for process-quality filtering and diagnosis, while treating $s_{\mathrm{ver}}$ as a separate outcome constraint.

\section{Rubric-Filtered SFT Details}
\label{app:sft-training}

We use the evolved rubrics as an offline process-quality filter before supervised fine-tuning. This choice isolates the effect of self-evolved rubrics as data selectors: the experiment asks whether the rubric can identify reusable skill-use demonstrations, rather than introducing additional optimization choices from on-policy RL. The same rubric-based meta score is compatible with RL as a process reward, since it provides dimension-level feedback on skill selection, key-step following, skill composition, and skill-grounded reflection. We leave RL optimization with \textsc{SkillCoach} rewards to future work.

We fine-tune Qwen3.5-4B and Qwen3.5-9B with full-parameter SFT using LLaMA-Factory. The training data include six variants: \texttt{full\_$R^{0}$}, \texttt{full\_$R^{\mathrm{best}}$}, \texttt{outcome\_only}, \texttt{$R^{\mathrm{best}}$\_minus\_order}, \texttt{$R^{\mathrm{best}}$\_minus\_reflect}, and \texttt{$R^{\mathrm{best}}$\_minus\_step}. Each variant is trained separately for the 4B and 9B base models, resulting in 12 SFT models. Here, skill selection is treated as a prerequisite rather than a removable scoring term.

The raw data are agent trajectories. The validation split is used only for checkpoint selection. We train for 5 epochs with learning rate $5.0\times10^{-6}$, cosine scheduling, warmup ratio 0.1, per-device batch size 1, gradient accumulation 1, bf16 precision, and DeepSpeed ZeRO-2. The final checkpoint for each run is selected by the lowest validation loss.

\tcbset{
  rubricprompt/.style={
    enhanced,
    breakable,
    colback=white,
    colframe=rubricblue,
    boxrule=0.8pt,
    arc=2pt,
    left=4pt,
    right=4pt,
    top=10pt,
    bottom=4pt,
    before skip=6pt,
    after skip=6pt,
    coltitle=white,
    fonttitle=\bfseries\footnotesize,
    attach boxed title to top left={xshift=6pt,yshift=-2pt},
    boxed title style={
      colback=rubricblue,
      colframe=rubricblue,
      boxrule=0pt,
      arc=2pt,
      left=5pt,
      right=5pt,
      top=2pt,
      bottom=2pt
    }
  }
}

\newtcblisting{promptbox}[1]{
  rubricprompt,
  title={#1},
  listing only,
  listing engine=listings,
  listing options={
    basicstyle=\ttfamily\scriptsize,
    breaklines=true,
    columns=fullflexible,
    keepspaces=true,
    showstringspaces=false
  }
}

\section{Implementation Details and Prompt Templates for Self-Evolving Rubrics}
\label{app:prompt-templates}

This appendix provides implementation details and near-complete prompt templates for the self-evolving rubric pipeline. We redact only task-specific payloads, full trajectory bodies, skill document contents, and service-specific metadata. Redacted fields are shown as \texttt{<... redacted ...>}. The hard verifier and the validation gate are implemented as code-level components, not LLM prompts.

\paragraph{Evolution protocol.}
For each task, rubric evolution is performed over a fixed pool of real execution trajectories. At each round, \textsc{SkillCoach} uses 10 trajectories as the calibration set $\mathcal{C}_t$ for judging and patch proposal, and uses a held-out validation set $\mathcal{U}_t$ to decide whether the candidate rubric should be accepted. When enough valid trajectories are available, the validation set contains 5 trajectories. Validation trajectories are hidden from the arbitration model.

The pipeline starts by extracting observable key steps from the task instruction, gold skills, oracle solution, verifier metadata, and sample rollouts. These key steps are then used to generate the initial rubric $R_t^0$. At round $j$, the current rubric $R_t^j$ judges the calibration trajectories, and the arbitration model proposes a localized patch $\Delta_t^j$ using only calibration evidence. Applying this patch gives a candidate rubric $\widetilde{R}_t^{j+1}$. The candidate and the current rubric are then evaluated on the same held-out validation set $\mathcal{U}_t$.

\paragraph{Validation gate.}
A candidate rubric is accepted only when it satisfies four conditions. First, it must not regress on the hard gate $H$, which prevents destructive changes such as unsupported skill-selection credit, unsupported key-step completion, ignored distractors, removed critical key steps, or verifier bypassing. Second, it must improve the soft objective $Q$ by more than $\epsilon=0.2$. Third, the improvement must include at least one material validation signal, rather than only increased judge confidence. Material signals include improved key-step evidence coverage, evidence quality, reflection evidence quality, process-verifier consistency, reduced hard-violation count, or a positive average score change in at least one process dimension. We use a component-level improvement threshold of 0.02 and a dimension-score improvement threshold of 0.01. Fourth, the patch must contain no structural violation, such as editing key steps, modifying score weights, deleting process dimensions, or altering the external verifier signal.

If the candidate passes the validation gate, it becomes the next accepted rubric version; otherwise, \textsc{SkillCoach} keeps the current rubric and continues to the next round. The final rubric $R_t^{\mathrm{best}}$ is selected from the accepted version set by validation performance, rather than simply taking the last generated version.

\paragraph{Stopping rule and run statistics.}
We set the maximum number of evolution rounds to $K_{\max}=6$. Evolution stops early if three consecutive candidates are rejected, if the arbitration model returns no usable patch, or if the patch is illegal or empty. The rejection counter is reset whenever a candidate patch is accepted. Across 28 tasks, the full run used 94 total evolution rounds, with an average of 3.36 rounds per task, a minimum of 3 rounds, and a maximum of 6 rounds. The prompt templates below, together with the validation-gate rules above, are sufficient to reproduce the self-evolving rubric pipeline.

\begin{promptbox}{Shared Ontology Inserted into LLM Prompts}
The four dimensions of agentic skill use (evaluate in this gating order):

1. skill_selection (GATE): did the agent select the required gold skill(s) and
   avoid distractor skills? In a no-gold setting, did it correctly refuse to use
   a skill? If this fails, downstream dimensions are discounted.
2. skill_following: did the agent actually perform the skill's KEY STEPS (not
   just name the skill)? Steps marked "not needed" for this instance do not
   count against coverage.
3. skill_composition_order: for multi-skill / multi-step tasks, are the step
   ORDER and the passing of intermediate artifacts between skills correct? If
   the task has a single gold skill this dimension is not_applicable.
4. result_reflection: before finishing, did the agent do an EXPLICIT, visible
   self-check / verification / reflection of its result? Only visible behavior
   counts; never assume hidden reasoning.
verifier: the task's hard verifier result. This is an external outcome signal produced by a rule runner, not an LLM judgment, and it is not part of the process meta score.
\end{promptbox}

\begin{promptbox}{Prompt 1: Key-Step Extraction}
SYSTEM PROMPT

You extract observable KEY STEPS for the skill_following dimension.

Your goal is not to summarize the task. Your goal is to define concrete,
evidence-checkable steps that a judge can later match against an agent
trajectory.

Use:
- the task instruction,
- the full gold skill content,
- the oracle solve.sh,
- the verifier description,
- sample real rollouts only for trajectory format and common error patterns.

The sample rollouts are NOT labels.

A valid key step must be observable in a trajectory. Do not write vague steps
such as:
- understand the task
- analyze the data
- use the skill
- produce the answer
- follow the workflow

Instead, write concrete steps such as:
- read a specific input file
- extract a station list
- compute a threshold table
- write a required output artifact
- run a verification command
- transform data from one schema to another

Output STRICT JSON ONLY:

{
  "key_steps": [
    {
      "step_id": "S1",
      "skill_id": "gold skill name or general",
      "description": "a concrete observable action",
      "critical": true,
      "source": "skill|oracle|verifier|rollout",
      "source_quote": "short quote from skill/oracle/verifier that supports this step",
      "evidence_type": "read|edit|execute|message|artifact|mixed",
      "positive_evidence": [
        "what trajectory evidence would show this step was completed"
      ],
      "negative_evidence": [
        "what trajectory evidence would show this step was skipped or done incorrectly"
      ],
      "required_before": [],
      "required_after": [],
      "expected_artifact": "file, table, variable, or intermediate result expected from this step; empty string if none",
      "optional_condition": "when this step may be marked not_needed; empty string if always needed",
      "failure_if_missing": "what error is likely if this step is missing",
      "score_impact": "critical_missing_caps_following_to_0.7|normal"
    }
  ]
}

Requirements:
1. Return at least 4 key steps unless the task is truly trivial.
2. At least one key step must come from the gold skill content.
3. At least one key step must be tied to the final artifact or verifier requirement.
4. Every critical step must include positive_evidence and negative_evidence.
5. Each description must describe an action that can be checked against tool
   calls, messages, files, commands, or artifacts.
6. Do not infer hidden reasoning. Only visible trajectory evidence counts.

Return ONLY the JSON object.

USER PROMPT TEMPLATE

Extract evidence-checkable key steps for task `<task_id>`.

<TASK_INSTRUCTION>
<task instruction redacted>
</TASK_INSTRUCTION>

<FULL_GOLD_SKILL_PACK::<skill_name>>
<full gold SKILL.md content redacted>
</FULL_GOLD_SKILL_PACK::<skill_name>>

<ORACLE_SOLUTION_solve_sh>
<oracle solution redacted>
</ORACLE_SOLUTION_solve_sh>

<TASK_VERIFIER>
<verifier code or verifier description redacted>
</TASK_VERIFIER>

<SAMPLE_REAL_ROLLOUT_1_for_format_and_common_errors_only>
<compact real rollout JSON redacted>
</SAMPLE_REAL_ROLLOUT_1_for_format_and_common_errors_only>

<SAMPLE_REAL_ROLLOUT_2_for_format_and_common_errors_only>
<compact real rollout JSON redacted>
</SAMPLE_REAL_ROLLOUT_2_for_format_and_common_errors_only>

<SAMPLE_REAL_ROLLOUT_3_for_format_and_common_errors_only>
<compact real rollout JSON redacted>
</SAMPLE_REAL_ROLLOUT_3_for_format_and_common_errors_only>

These three sample rollouts are separate examples for trajectory format and
common error patterns. They are not labels.

Return ONLY the JSON object.
\end{promptbox}

\begin{promptbox}{Prompt 2: Initial Rubric Generation R0}
SYSTEM PROMPT

You are a meticulous evaluation-rubric author for agentic skill use.

Produce an initial task-level rubric R0 for judging real agent trajectories.

Important:
- The key steps have already been extracted.
- You must NOT invent a second set of key steps.
- The field dimensions.skill_following.key_steps must exactly use the provided
  EXTRACTED_KEY_STEPS.
- You may write criteria, evidence requirements, negative cases, and score rules
  around those key steps.

[SHARED ONTOLOGY INSERTED HERE]

Output STRICT JSON ONLY, exactly this shape:

{
  "rubric_id": "<task_id>_R0",
  "task_id": "<task_id>",
  "version": 0,
  "created_by": "gpt_5_5",
  "dimensions": {
    "skill_selection": {
      "criteria": [],
      "negative_cases": [],
      "evidence_requirements": [],
      "score_rules": []
    },
    "skill_following": {
      "criteria": [],
      "key_steps": [],
      "evidence_requirements": [],
      "score_rules": []
    },
    "skill_composition_order": {
      "applicable": true,
      "expected_order": [],
      "dependencies": [],
      "handoff_requirements": [],
      "criteria": [],
      "score_rules": []
    },
    "result_reflection": {
      "criteria": [],
      "evidence_requirements": [],
      "score_rules": []
    },
    "verifier": {
      "criteria": ["Use the hard verifier result only."],
      "score_rules": []
    }
  },
  "score_weights": {
    "skill_selection": 0.40,
    "skill_following": 0.30,
    "skill_composition_order": 0.20,
    "result_reflection": 0.10
  }
}

Requirements:
1. skill_selection must name the actual gold skills and distractor skills.
2. skill_selection must distinguish real skill use (SKILL.md was read) from
   merely mentioning a skill name.
3. skill_following.key_steps must be exactly the provided EXTRACTED_KEY_STEPS.
4. skill_following criteria and score_rules must refer to key step IDs.
5. If there is only one gold skill, set skill_composition_order.applicable to false.
6. If there are multiple gold skills or ordered substeps, fill expected_order,
   dependencies, and handoff_requirements.
7. result_reflection only counts visible self-checking behavior.
8. verifier is not judged by an LLM; it comes from the hard benchmark verifier.
9. Sample real rollouts are only for trajectory format and common mistakes. They
   are not labels.

Return ONLY the JSON rubric.

USER PROMPT TEMPLATE

Generate the R0 rubric for task `<task_id>`.

<TASK_INSTRUCTION>
<task instruction redacted>
</TASK_INSTRUCTION>

<TASK_PACKAGE>
<task package JSON redacted>
</TASK_PACKAGE>

<GOLD_SKILL::<skill_name>>
<full gold SKILL.md content redacted>
</GOLD_SKILL::<skill_name>>

<DISTRACTOR_SKILL::<skill_name>>
<distractor SKILL.md content redacted>
</DISTRACTOR_SKILL::<skill_name>>

<ORACLE_SOLUTION_solve_sh>
<oracle solution redacted>
</ORACLE_SOLUTION_solve_sh>

<TASK_VERIFIER>
<verifier code or verifier description redacted>
</TASK_VERIFIER>

<EXTRACTED_KEY_STEPS>
<key step JSON from Prompt 1>
</EXTRACTED_KEY_STEPS>

<SAMPLE_REAL_ROLLOUT_1_for_format_and_common_errors_only>
<compact real rollout JSON redacted>
</SAMPLE_REAL_ROLLOUT_1_for_format_and_common_errors_only>

<SAMPLE_REAL_ROLLOUT_2_for_format_and_common_errors_only>
<compact real rollout JSON redacted>
</SAMPLE_REAL_ROLLOUT_2_for_format_and_common_errors_only>

<SAMPLE_REAL_ROLLOUT_3_for_format_and_common_errors_only>
<compact real rollout JSON redacted>
</SAMPLE_REAL_ROLLOUT_3_for_format_and_common_errors_only>

These three sample rollouts are separate examples for trajectory format and
common mistakes; they are NOT labels.

Return ONLY the JSON rubric. Do not invent additional key_steps. Use
EXTRACTED_KEY_STEPS exactly.
\end{promptbox}

\begin{promptbox}{Prompt 3: Skill Selection Evidence Boundary}
IMPLEMENTATION NOTE

Current implementation is rule-dominant for skill_selection. Kimi is not called
for this dimension. The following is the complete evidence-boundary
specification used to separate skill selection evidence from method evidence.
The same boundary is also reflected in the optional LLM-format judge prompt kept
in the prompt library.

RULE SPECIFICATION

Decide whether the agent selected and used the required gold skill(s), and
whether it incorrectly used distractor skills.

Use mechanical evidence first:
- gold_skill_doc_read and distractor_doc_read are selection evidence.
- Reading SKILL.md or a visible "Launching skill:" event counts as skill
  selection.
- pypdf, fitz, PdfReader, pandas, csv/pdf filenames, or other library/file usage
  is method evidence only.
- method evidence must not be treated as selecting a skill.

A skill is selected when the agent READs its SKILL.md. Mentioning a skill name
without reading it is NOT selection; that is a skill_following issue.

OUTPUT OBJECT

{
  "dimension": "skill_selection",
  "score": 0.0,
  "label": "correct|partial|wrong|missing",
  "gold_skill_used": false,
  "distractor_used": false,
  "false_trigger": false,
  "evidence": [
    {
      "span": "short trajectory evidence",
      "reason": "why this supports the selection judgment"
    }
  ],
  "failure_reason": "",
  "confidence": 0.95,
  "selection_evidence_types": {
    "skill_doc_read": false,
    "gold_method_evidence": false,
    "distractor_doc_read": false,
    "distractor_method_use": false,
    "method_evidence": false
  },
  "llm_assist": {
    "used": false,
    "reason": "skill_selection is rule-dominant; Kimi is not called for this dimension"
  }
}

RULES

1. correct means all required gold skills were selected and no harmful
   distractor was used.
2. partial means a gold skill was read or invoked, but distractor evidence also
   appears.
3. wrong means the agent mainly selected a distractor or used the wrong skill
   path.
4. missing means no gold skill evidence is found.
5. false_trigger is true when the agent uses a skill in a no-gold setting or
   forces an irrelevant skill.
6. Every positive judgment must cite event_index evidence.

INPUT TEMPLATE

<EVENT_INDEXED_TIMELINE>
<compact timeline with event_index retained>
</EVENT_INDEXED_TIMELINE>

<SKILL_EVENTS>
<skill event JSON, if present>
</SKILL_EVENTS>

<GOLD_SKILLS>
<gold skill names>
</GOLD_SKILLS>

<DISTRACTOR_SKILLS>
<distractor skill names>
</DISTRACTOR_SKILLS>
\end{promptbox}

\begin{promptbox}{Prompt 4: Skill Following Judge}
SYSTEM PROMPT

You are the SKILL_FOLLOWING judge.

Your job is to decide whether the agent performed each extracted key step using
visible trajectory evidence.

You must NOT judge hidden reasoning.
You must NOT give credit for merely mentioning a skill name.
You must NOT mark a step completed without concrete evidence.
You should first inspect DETECTED_EVIDENCE.key_step_evidence, then verify the
cited event_index in EVENT_INDEXED_TIMELINE.
DETECTED_EVIDENCE is candidate evidence only; irrelevant matches do not count.
Keep any internal analysis brief. Always finish with the strict JSON object.

[SHARED ONTOLOGY INSERTED HERE]

For each key step, assign one status:
- completed: clear trajectory evidence shows the step was done correctly
- partial: some evidence exists, but the step is incomplete or shallow
- missing: no evidence that the step was done
- wrong: the agent attempted the step but did it incorrectly
- not_needed: the key step is genuinely unnecessary for this instance,
  according to optional_condition

Evidence must cite event_index from EVENT_INDEXED_TIMELINE.

Output STRICT JSON ONLY:

{
  "dimension": "skill_following",
  "key_steps": [
    {
      "step_id": "S1",
      "status": "completed|partial|missing|wrong|not_needed",
      "evidence": [
        {
          "event_index": 0,
          "span": "short quoted evidence from the trajectory",
          "reason": "why this evidence supports the status"
        }
      ],
      "comment": "short explanation"
    }
  ],
  "failure_reason": "",
  "confidence": 0.0
}

Rules:
1. Do not output score.
2. Do not output critical_step_coverage.
3. The code will compute score and coverage later.
4. completed and partial require at least one event_index evidence item.
5. missing may have empty evidence.
6. not_needed must cite the key step's optional_condition.
7. If there is no gold skill invocation evidence, do not mark critical
   skill-specific steps as completed.

Return ONLY that JSON object.

USER PROMPT TEMPLATE

Judge dimension `skill_following` for a trajectory on task `<task_id>`.

<DIMENSION_RUBRIC>
<skill_following rubric JSON, including key_steps>
</DIMENSION_RUBRIC>

<GOLD_SKILLS>
<gold skill names>
</GOLD_SKILLS>

<DISTRACTOR_SKILLS>
<distractor skill names>
</DISTRACTOR_SKILLS>

<DETECTED_EVIDENCE>
{
  "schema_version": 1,
  "selection_evidence_summary": {
    "gold_skill_doc_read": [],
    "gold_method_evidence": [],
    "distractor_doc_read": [],
    "distractor_method_use": []
  },
  "key_step_evidence": [
    {
      "step_id": "S1",
      "candidates": [
        {
          "event_index": 0,
          "evidence_span": "candidate evidence span",
          "evidence_kind": "read|edit|execute|message|artifact|mixed"
        }
      ]
    }
  ],
  "note": "deterministic candidate evidence only"
}
</DETECTED_EVIDENCE>

DETECTED_EVIDENCE contains deterministic candidate evidence only. Use it to
locate visible support, but do not treat it as automatic credit.

<EVENT_INDEXED_TIMELINE>
<focused compact timeline JSON with event_index retained>
</EVENT_INDEXED_TIMELINE>

<SKILL_EVENTS>
<skill events JSON>
</SKILL_EVENTS>

<VERIFIER_RESULT>
<hard verifier result JSON>
</VERIFIER_RESULT>

Use only visible trajectory evidence. Cite event_index whenever possible. Return
ONLY the JSON object for this dimension.
\end{promptbox}

\begin{promptbox}{Prompt 5: Composition Order Judge}
SYSTEM PROMPT

You are the SKILL_COMPOSITION_ORDER judge.

Judge whether the agent used multiple skills or key steps in the correct order,
and whether intermediate artifacts were passed correctly between them.

If the rubric says applicable=false, return score 1.0 and order_correct=null.

[SHARED ONTOLOGY INSERTED HERE]

Use:
- expected_order
- dependencies
- handoff_requirements
- event-indexed trajectory evidence

Output STRICT JSON ONLY:

{
  "dimension": "skill_composition_order",
  "score": 0.0,
  "expected_order": [],
  "observed_order": [],
  "order_correct": true,
  "composition_errors": [
    {
      "type": "wrong_order|missing_dependency|missing_handoff|redundant_skill",
      "event_index": 0,
      "evidence": "short trajectory evidence",
      "comment": "why this is a composition error"
    }
  ],
  "confidence": 0.0
}

Rules:
1. First infer observed_order from the trajectory.
2. Compare observed_order with expected_order.
3. Check whether each dependency's artifact was produced before it was consumed.
4. Check whether handoff_requirements are satisfied.
5. If there is only one gold skill and no ordered dependencies, return score 1.0
   and order_correct=null.
6. Cite event_index evidence for every error.

Return ONLY that JSON object.

USER PROMPT TEMPLATE

Judge dimension `skill_composition_order` for a trajectory on task `<task_id>`.

<DIMENSION_RUBRIC>
<composition rubric JSON with expected_order, dependencies, and handoff requirements>
</DIMENSION_RUBRIC>

<GOLD_SKILLS>
<gold skill names>
</GOLD_SKILLS>

<DISTRACTOR_SKILLS>
<distractor skill names>
</DISTRACTOR_SKILLS>

<EVENT_INDEXED_TIMELINE>
<compact event-indexed timeline JSON>
</EVENT_INDEXED_TIMELINE>

<SKILL_EVENTS>
<skill events JSON>
</SKILL_EVENTS>

<VERIFIER_RESULT>
<hard verifier result JSON>
</VERIFIER_RESULT>

Use only visible trajectory evidence. Cite event_index whenever possible. Return
ONLY the JSON object for this dimension.
\end{promptbox}

\begin{promptbox}{Prompt 6: Result Reflection Judge}
SYSTEM PROMPT

You are the RESULT_REFLECTION judge. Decide whether the agent did an EXPLICIT,
visible self-check / verification / reflection of its result before finishing.
Only visible trajectory behavior counts; never assume hidden reasoning. Absence
of reflection should NOT, by itself, make the whole trajectory a failure.

[SHARED ONTOLOGY INSERTED HERE]

Output STRICT JSON ONLY:

{
  "dimension": "result_reflection",
  "score": 0.0,
  "has_explicit_check": false,
  "check_quality": "strong|weak|invalid|missing",
  "reflection_evidence": [
    {
      "event_index": 0,
      "span": "",
      "reason": ""
    }
  ],
  "missed_errors": [],
  "confidence": 0.0
}

Return ONLY that JSON object.

USER PROMPT TEMPLATE

Judge dimension `result_reflection` for a trajectory on task `<task_id>`.

<DIMENSION_RUBRIC>
<reflection rubric JSON>
</DIMENSION_RUBRIC>

<GOLD_SKILLS>
<gold skill names>
</GOLD_SKILLS>

<DISTRACTOR_SKILLS>
<distractor skill names>
</DISTRACTOR_SKILLS>

<EVENT_INDEXED_TIMELINE>
<compact event-indexed timeline JSON>
</EVENT_INDEXED_TIMELINE>

<SKILL_EVENTS>
<skill events JSON>
</SKILL_EVENTS>

<VERIFIER_RESULT>
<hard verifier result JSON>
</VERIFIER_RESULT>

Use only visible trajectory evidence. Cite event_index whenever possible. Return
ONLY the JSON object for this dimension.
\end{promptbox}

\begin{promptbox}{Prompt 7: Rubric Patch Arbiter}
SYSTEM PROMPT

You are the rubric ARBITER for real rollout rubric evolution.

You receive ONLY calibration rollout evidence. You must not assume or request
any validation rollout evidence.

Your job is to propose one SMALL structured patch to the current rubric. Never
rewrite the full rubric.

Hard constraints:
- Do NOT edit dimensions.skill_following.key_steps.
- Do NOT delete dimensions.
- Do NOT modify or bypass the external verifier signal.
- Do NOT remove gold skill information from skill_selection.
- Do NOT change score_weights.
- Only patch criteria, negative_cases, evidence_requirements, score_rules,
  composition dependencies, expected_order, or handoff_requirements.
- If there is no useful localized patch, return {"edits":[]}.

[SHARED ONTOLOGY INSERTED HERE]

Output STRICT JSON ONLY:

{
  "patch_id": "",
  "from_version": <j>,
  "to_candidate_version": <j+1>,
  "edits": [
    {
      "operation": "<op>",
      "dimension": "<one of the four>",
      "target_path": "dimensions.<dim>.<allowed_list>",
      "content": {...},
      "reason": "calibration rollout/judge evidence motivating this edit",
      "supporting_trajectory_ids": ["..."]
    }
  ],
  "expected_improvement": ["..."],
  "risk": ["..."]
}

Allowed operations:
add_criterion, delete_criterion, replace_criterion, tighten_condition,
relax_condition, add_negative_case, add_evidence_requirement,
add_uncertainty_rule, add_score_rule, modify_score_rule, set_expected_order,
add_dependency, add_handoff_requirement

ID discipline:
- PATCHABLE_TARGETS lists the only existing ids you may modify or delete.
- delete_criterion, replace_criterion, modify_score_rule, tighten_condition, and
  relax_condition MUST use an id from PATCHABLE_TARGETS.existing_ids.
- Never invent a modify/delete id.
- To add a new score rule, use add_score_rule. Do not use modify_score_rule to
  add a new rule.

Composition operation content:
- add_score_rule: target_path = "dimensions.<dim>.score_rules",
  content = {"id":"","text":""}
- set_expected_order: content = {"expected_order":["step or skill id", "..."]}
- add_dependency: target_path = "dimensions.skill_composition_order.dependencies",
  content = {"id":"","from":"","to":"","artifact":"","text":""}
- add_handoff_requirement:
  target_path = "dimensions.skill_composition_order.handoff_requirements",
  content = {"id":"","from":"","to":"","artifact":"","text":""}

Return ONLY one JSON patch object.

USER PROMPT TEMPLATE

Propose ONE localized patch to improve the rubric for task `<task_id>` using
calibration rollouts only.

<TASK_PACKAGE>
<task package JSON redacted>
</TASK_PACKAGE>

<CURRENT_RUBRIC>
<current rubric JSON R_j>
</CURRENT_RUBRIC>

<CALIBRATION_JUDGE_RESULTS>
<per-rollout, per-dimension judge results on calibration rollouts>
</CALIBRATION_JUDGE_RESULTS>

<CALIBRATION_REAL_ROLLOUT_TRAJECTORIES>
<up to three compact calibration rollout JSON objects redacted>
</CALIBRATION_REAL_ROLLOUT_TRAJECTORIES>

<MECHANICAL_SKILL_DETECTION>
<mechanical evidence for gold skill doc read, method evidence, distractor doc read,
and distractor method use>
</MECHANICAL_SKILL_DETECTION>

<PATCHABLE_TARGETS>
{
  "existing_ids": ["criterion_id", "score_rule_id", "..."],
  "by_dimension": {
    "skill_selection": ["..."],
    "skill_following": ["..."],
    "skill_composition_order": ["..."],
    "result_reflection": ["..."]
  }
}
</PATCHABLE_TARGETS>

<VERSION_HISTORY>
<accepted and rejected version history>
</VERSION_HISTORY>

<REJECTED_PATCHES_do_not_repeat>
<recent rejected patches>
</REJECTED_PATCHES_do_not_repeat>

Validation rollouts are hidden from you. Return ONLY one JSON patch object. Do
not modify key_steps.
\end{promptbox}

\begin{promptbox}{Non-Prompt Components}
These components are part of the method, but they are not LLM prompts.

HARD VERIFIER

The verifier dimension is not judged by Kimi or GPT-5.5. It only reads benchmark
outputs such as reward.txt, result.json rewards, or test-stdout.txt. The arbiter
cannot change these results.

VALIDATION GATE

Patch acceptance is a code-level decision. A candidate rubric must not regress
on hard gate score, and its soft quality must improve by a threshold with
material evidence improvement. Material improvements may include:
- key_step_evidence_coverage improvement
- evidence_quality improvement
- reflection_evidence_quality improvement
- process_verifier_consistency improvement
- hard violation count reduction
- positive dimension score change

Improvements from judge confidence alone are rejected.

SCORE AGGREGATION

For skill_following, Kimi outputs step statuses and the code aggregates them.
Critical steps receive higher weight. completed and partial require event_index
evidence; otherwise they are downgraded. Empty support sets are not treated as
perfect evidence. The report records support_counts and uninformative_components.
\end{promptbox}

\newpage

\end{document}